\documentclass[runningheads]{llncs}
\usepackage[T1]{fontenc}
\usepackage[pages=all, color=gray, opacity=0.2, scale=3, angle=45]{background}
\backgroundsetup{
  contents={PREPRINT}
}
\usepackage{graphicx}
\usepackage{amsmath}
\usepackage{amssymb}
\usepackage{comment}
\usepackage{xcolor}
\usepackage{booktabs}
\usepackage{float}
\usepackage{enumitem}
\usepackage{caption}
\usepackage[table]{xcolor}
\usepackage[subtle]{savetrees}
\usepackage{wrapfig}
\usepackage{placeins}
\newcounter{todocounter}

\begin{document}
\title{GNN4PPM: Multi-Target Predictive Process Monitoring with Relational Graph Convolutional Networks}
\titlerunning{GNN4PPM}
% If the paper title is too long for the running head, you can set
% an abbreviated paper title here
%
%\author{Anonymous Author(s)}
\author{\mbox{Ana Costa\inst{1}\orcidID{0000-0001-9241-5614}} \and
\mbox{Johannes Mäkelburg\inst{1}\orcidID{0009-0001-3821-7817}} \and
\mbox{Luise Pufahl\inst{1}\orcidID{0000-0002-5182-2587}}}

%\authorrunning{Author et al.}
\authorrunning{A. Costa et al.}

%\institute{Anonymous Institution(s)}
\institute{Technical University of Munich, School of Computation, Information and Technology \\ 
Heilbronn, Germany \\
\email{a.costa@tum.de, johannes.maekelburg@tum.de, luise.pufahl@tum.de}\\
%\url{https://www.cs.cit.tum.de/infs/startseite/}
}
\maketitle              % typeset the header of the contribution
\begin{abstract}

Predictive Process Monitoring (PPM) aims at predicting at runtime and as early as possible the future states of a process execution. Common tasks include predicting the next event, the time to completion of a trace, and outcomes.
Existing approaches typically consider an event from the perspective of the executed activities along with their timestamps and case identifiers. 
This leads to the disadvantage that in real-life settings, there is much more information recorded in the event log that is not captured or completely ignored when performing prediction tasks. 
We introduce GNN4PPM, an approach that predicts all next events along with their complete data payload at once. 
We represent event information in a heterogeneous knowledge graph that captures the event log as an RDF semantics, and train the embeddings with a Relational Graph Convolutional Network (R-GCN). 
Our approach is promising in comparison to existing solutions, and experiments with state-of-the-art solutions prove the accuracy and applicability of GNN4PPM in complex settings.

\begin{comment}
- Intro to PPM and GNN, what is PPM, how does GNN architecture helps? \newline 
- Problem with multi-attribute prediction \newline
- Solution with temporal GNN architecture \newline
- Model/ evaluation/ results
\end{comment}

\end{abstract}
\section{Introduction}
Predictive Process Monitoring (PPM) is a branch of process mining that focuses on forecasting the future of an ongoing business process case \cite{ceravolo_predictive_2024}.
Information from historical event data for each case is recorded, such as the order of executed activities, their timestamp, and additional attributes \cite{maggi_predictive_2014}. 
Instead of looking only at what has happened, predictive monitoring helps in understanding what will happen next, allowing process stakeholders to intervene before issues occur. 
PPM uses as input data an event log, a structured collection of event traces that record how individual process cases evolve \cite{diFrancescomarino2022predictive}. Each event refers to an executed activity and includes at least a case identifier and a timestamp, optionally enriched with event attributes
%that constitute the data payload, 
while traces themselves may contain case‑level attributes.

One of the tasks in PPM, the next event prediction, focuses on predicting sequences of future activities and related data payloads \cite{ceravolo_predictive_2024}. These data payloads refer to attributes associated with events, such as the timestamp or the resource performing the next activity.
A naive approach is to consider only the recorded activity within its case, along with its timestamps \cite{diFrancescomarino2022predictive}. 
%However, such predictive methods do not consider the context in which a case and its events are executed \cite{Camargo2019}. 
This leads to the disadvantage that in real-life settings, there is much extra \emph{information recorded in the event log that is not captured or is completely ignored when performing prediction tasks}, such as the resource of the event, its role, the characteristics of the case, and other relevant attributes \cite{Dissegna2025}. 
For example, in the loan application process at a financial institution, loan acceptance depends on a combination of application characteristics, such as application type, loan goal, requested amount, and other institutional context \cite{vandongen2017bpi}. Considering only the activity along with the timestamp ordering for prediction provides poor information on a loan acceptance or rejection.

Even though business processes are traditionally represented and discovered in process models as graph structures with their activities, execution relations, and case information~\cite{harl_explainable_2020}, predictive approaches do \emph{not properly capture the interrelations between events, case attributes, and event attribute}s. As a result, contextual information available in logs is not fully exploited during the prediction. 
%have been developed to extract sequential data from past process executions and use this knowledge for complex predictions. 
Many deep learning approaches do not take advantage of the dependencies between events since the prefix is embedded into a single vector. Therefore, features of each execution can be overwritten~\cite{rama-maneiro_embedding_2021}. 

%In addition to that, process experts usually collect extensive information to make a decision, and prediction analysts must limit this information by treating a single event attribute as the prediction target \cite{Camargo2019}. 
%As a result, the predictive model needs to be trained for each event attribute in the same event log several times \cite{tax_predictive_2017}. 
%This requires significant manual effort and training time that could be reduced if multiple attributes could be captured during the same training. 
In predictive process monitoring, the prediction target can vary across different event attributes (e.g., activity, resource, loan risk value), yet most approaches \emph{focus on a single target attribute} \cite{Camargo2019}. As a result, separate models must be trained from scratch for each attribute on the same event log \cite{tax_predictive_2017}, leading to increased manual effort and training time. When in reality, prediction models are so target-specific that even when using the same event log with a different target, the model needs to be completely retrained \cite{mehdiyev_interpretable_2025}.

We consider that event and case attributes should participate in triggering the future events and their attributes. 
We predict, therefore, in our multi-target model, all the next events along with the complete data payload of the event log at once. This enables the model to capture inter-attribute dependencies that are ignored by single-target approaches while reducing the retraining effort required for different prediction targets.
To incorporate the contextual relations among event attributes into the learning process \cite{weinzierl_exploring_2022} and the sequence aspects of the traces \cite{Wuyts2024}, our prediction model combines sequential information and graph embeddings into the Graph Neural Network (GNN) training. 
In principle, neighbours can be aggregated for more informative node embeddings \cite{agarwal_evaluating_2023} by using a Relational Graph Convolutional Network (R-GCN) architecture, which is designed to handle such heterogeneous graphs \cite{schlichtkrull2018rgcn}. 
The strong performance of this type of neural network on graph-structured data \cite{chen2019grgnn,schlichtkrull2018rgcn} and their ability to handle highly multi-relational data characteristics motivate our interest in exploring the architecture. 
Our contributions include, therefore (1) the design of a heterogeneous knowledge graph capable of including the complete data payload of the event log and its relations, (2) the combination of sequential information and RDF2Vec embeddings for representation of trace characteristics, and (3) the prediction of all next event attributes using an R-GCN architecture with focus on the attributes neighborhood relations.

The paper is organized as follows. Section~\ref{RW} and \ref{PRE} describe recent achievements in literature, preliminaries, and problem definition. Section~\ref{APP} describes our approach, section~\ref{EVA} describes the experimental setup and evaluation results, and finally, section~\ref{CON} presents the conclusion and future work.

\begin{comment}
Advantages:
The use of graphs enhances a higher degree of comprehension among domain experts, making the information more explainable \cite{agarwal_evaluating_2023}
GNN: In particular, the structure of the input graph can be matched directly to the topology of the GNN; this allows for direct inferences to be made between the relevance of network nodes and graph nodes \cite{harl_explainable_2020}.

Evaluation - predict other events besides the activity
For other approaches/ event logs, predict other attributes

What is done in the paper
How is the study conducted
The findings
Outline

- Reviewed by someone who is not into process prediction
- BPM expert, but not in details from Process Prediction
- Caise wants this relation to Information Sytems - CAISE topic is reflective to information systems architecture
- Process data coming to PPM
\end{comment}

\section{Related Work}\label{RW}

Although previous studies in PPM focus on exploring different model architectures for next event prediction \cite{Bukhsh2021,Rauch2026}, few approaches consider the prediction of more than one event attribute in the same model. \cite{Evermann2017} addresses the prediction of the next event, showing that Recurrent Neural Networks (RNNs) can capture sequential dependencies in event logs, but focusing on one event attribute. 
\cite{tax_predictive_2017} investigates Long Short-Term Memory (LSTMs) for the prediction of the next event and its timestamp, and \cite{Camargo2019} studies how to train RNNs with LSTMs to predict the next activity, its timestamp, and the associated resource as separate events. 
While \cite{Theis2019} focuses on predicting one attribute of the event by enhancing a discovered Petri net model with time decay functions, \cite{DiMauro2019} and \cite{pasquadibisceglie_using_2019} propose Convolutional Neural Network (CNN) architectures specifically for Next Activity Prediction (NAP). 
This architecture is further explored by other authors, who predict the next activity by enriching instance graphs \cite{chiorrini_multi-perspective_2023} or by adding an explainable layer after the prediction \cite{Aversano2023}.

Recent studies focus on suffix prediction mainly using activity as the event attribute \cite{Ali2023}. 
While \cite{Wuyts2024} addresses full-context-aware (i.e., considering also event attributes) suffix prediction, \cite{pasquadibisceglie_lupin_2024} proposes an LLM method to generate the sequence of future activities from a given prefix. 
Using a graph-based representation, \cite{pasquadibisceglie_prophet_2024} introduces an explainable approach that handles the next activity prediction, but \cite{Dissegna2025} integrates activities, timestamps, resources, and shows superiority in terms of accuracy compared to the previous method with a similar architecture. The few approaches that extend next event prediction to other attributes within the same model include, therefore, the timestamp of the activity or its resource \cite{Dissegna2025}. The reasoning behind this lack of event attribute consideration emphasizes that current deep learning architectures, such as RNNs and LSTMs, represent the events as sequential data rather than with the multi-target view that GNNs can capture.

Some authors, such as \cite{pasquadibisceglie_multi-view_2022,pasquadibisceglie_leveraging_2021}, recognize the problem by presenting multiple process perspectives in a multi-view setting, but target again, only the prediction of one event at a time. 
Other authors, such as \cite{BergerWolf2019}, extend next activity prediction to incorporate other attributes, but disregard time or numerical attributes. 
Since graph-based approaches facilitate the identification of the behavioral patterns \cite{rama-maneiro_embedding_2021}, some authors have investigated GNNs for both next activity prediction \cite{weinzierl_exploring_2022,chiorrini_multi-perspective_2023,venugopal_comparison_2021} and remaining time \cite{duong_remaining_2023,elyasi_pgtnet_2024}. Therefore, approaches are still limited to having as a target attribute one or a few attributes.

Several graph-based approaches target next-event prediction using homogeneous graphs \cite{weinzierl_exploring_2022,chiorrini_multi-perspective_2023,venugopal_comparison_2021}. Two approaches extend this architecture to heterogeneous graphs, which encode multiple node and edge types \cite{pasquadibisceglie_prophet_2024,Dissegna2025}. The advantage of using heterogeneous graphs is that they directly embrace the multi-target representation of event data \cite{pasquadibisceglie_prophet_2024}, which is what an event log with multiple event attributes contains. However, these previous approaches opt for Graph Attention Networks (GAT) instead of R-GCN. Since R-GCNs are developed specifically to handle the highly multi-relational data characteristic of realistic knowledge bases \cite{schlichtkrull2018rgcn}, we scale the relational data of our graphs by applying this neural architecture, which, to our knowledge, has not been explored before for the multi-target prediction problem within PPM.

\section{Preliminaries}\label{PRE}
 
This section describes the essential definitions used in the approach, as well as the problem statement. We first introduce the formal definitions of the key data constructs of PPM, such as events, traces, prefixes, and logs, drawing on established literature \cite{guizzardi_enhancing_2024}.

\begin{definition}[Event]
\label{def:event}
An \emph{event} is a tuple 
$e = (a, \mathit{cid}, t, v_1, \ldots, v_n)$ where $a \in A$ represents the activity label, $\mathit{cid} \in C$ is the unique identifier for the case, also called \emph{case ID}, $t \in \mathcal{T}$ represents the event’s timestamp and $v_1, \ldots, v_n$ are the event-specific attributes, where 
$\forall\, 1 \leq k \leq n : v_k \in \mathcal{V}_k$ is the domain of the $k^{th}$ attribute. These variables create a multi-dimensional space for the universe of events $\mathcal{E}$.
\end{definition}
\begin{definition}[Trace]
A \emph{trace} $\sigma \in \mathcal{E}^*$ is a finite sequence of unique events $\sigma = \langle e_1, e_2, \ldots, e_{|\sigma|} \rangle,$
where $|\sigma|$ indicates the number of events in the trace, also called the \emph{trace length}.  
The events are ordered chronologically and share the case identifier $\mathit{cid}$.  
\end{definition}

Similarly to the \emph{event-specific attributes}, each trace contains a set of \emph{trace-specific attributes}, which remain constant throughout the entire trace.  
Every trace mandatorily includes the \emph{case ID}, and may optionally include trace-specific attributes 
$w_1, \ldots, w_m$, where each $w_j \in \mathcal{W}_j$ belongs to the domain of the $j^{th}$ attribute. We define the set of all traces by $\mathcal{S} \subseteq \mathcal{E}^*$, with each trace $\sigma \in \mathcal{S}$. 

\begin{definition}[Prefix and Suffix]
The \emph{prefix} and \emph{suffix} are specific types of partial traces, obtained by employing the $hd^i(\sigma)$ and $tl^i(\sigma)$ functions, respectively.  This is realized with a selection operator $(\cdot): \sigma(i) = e_i, \forall i \in [1, |\sigma|] \subset \mathbb{N}$, such that
$hd^i(\sigma) = \langle e_1, e_2, \ldots, e_{\min(i, |\sigma|)} \rangle
\quad \text{and} \quad
tl^i(\sigma) = \langle e_y, e_{y+1}, \ldots, e_{|\sigma|} \rangle,
$ where $y = \max(1, |\sigma| - i + 1)$.
\end{definition}

\begin{definition}[Event Log] An \emph{event log} $L$ is a set of traces denoted by $L = \{ \sigma_1, \sigma_2, \ldots, \sigma_z \}$ and $\sigma_i \in \mathcal{S}$ for $1 \le i \le z$, $z \in \mathbb{N}^+$. Each $\sigma_i$ is a trace as previously defined. %The event log $L$ is a collection of traces that share the same unique identifiers $\mathit{cid} \in C$. 
\end{definition}

In this work, we make use of knowledge graphs derived from an event log. 
%For this purpose, we introduce the concept of knowledge graphs~\cite{hogan2021knowledge}. 
Following the Resource Description Framework (RDF) standard~\cite{rdfspecs},  a knowledge graph (KG)~\cite{hogan2021knowledge} is a collection of  \textit{(subject, predicate, object)} triples. 
Subjects and predicates are Internationalized Resource Identifiers (IRIs), while objects can either be IRIs representing entities or literals with a specific datatype (e.g., strings or integers).

\begin{definition}[Knowledge Graph]
A knowledge graph is a set of triples
$\mathcal{KG} \subseteq (\text{IRI} \times \text{IRI} \times (\text{IRI} \cup \text{Literal}))$,
where each triple $(s,p,o) \in \mathcal{KG}$ consists of a subject $s$, 
a predicate $p$, and an object $o$.
\begin{comment}
A Knowledge Graph $\mathcal{KG}$ is denoted by a tuple $(\mathbb{E}, \mathcal{R}, \mathcal{L}, \mathcal{C})$, where the pair-wise disjoint sets $\mathbb{E}$, $\mathcal{R}$, $\mathcal{L}$, and $\mathcal{C}$ correspond to the set of entities, relations, literals, and types or classes, respectively. A statement in $\mathcal{KG}$ is modelled as a triple $(s,r,o)$, with $s \in \mathbb{E} \cup \mathcal{C}$, $r \in \mathcal{R}$, and $o \in \mathbb{E} \cup \mathcal{L} \cup \mathcal{C}$.    
\end{comment}
\end{definition}
\vspace{-0.2cm}

In the context of logs, events, cases, and categorical attributes are represented as entities, whereas numerical values, such as timestamps, are represented as literals.

\vspace{-0.3cm}
\setlength{\parindent}{0pt}
\subsubsection*{Problem definition}
\label{def:problem}
Given an event log $L = \{ \sigma_1, \sigma_2, \ldots, \sigma_z \}$
%of completed traces 
where each trace
$\sigma = \langle e_1, e_2, \ldots, e_{|\sigma|} \rangle$
consists of events $e_i= (a_i,\mathit{cid}_i,t_i,v_{1_i}, \ldots, v_{n_i})$ %as defined in Definition~\ref{def:event}, 
%which consists of a collection of events $e_1, \ldots, e_{|\sigma|} $ 
%with $e_j= (a_j,c_j,t_j,v_{1_j}, \ldots, v_{n_j})$ 
%and $v_1, \ldots, v_n$ are the event-specific attributes, where $\forall\, 1 \leq i \leq n : v_i \in \mathcal{V}_i$ is the domain of the $i^{th}$ attribute
, and an event prefix $hd^i(\sigma)$ with $i < |\sigma|$. 
The task is to learn a predictive function 
$f: hd^i(\sigma) \rightarrow \hat{e}_{i+1} $ that predicts the next event $\hat{e}_{i+1}$,
such that $\hat{e}_{i+1}$ approximates the ground-truth $e_{i+1}$. 
The prediction target includes the activity label, timestamp, and all event-specific attributes.

%$f_{v}(Log, \sigma_i^{|\sigma|})$ 
%that returns the next event $e_{|\sigma| +1}$ that is as close as possible to to $v_{{|\sigma|}+1}$ %$, \ldots, n$
%, so that the output of the predictive model is the next event along with its attributes.

\section{GNN4PPM}\label{APP}

\begin{comment}
Pipeline of the steps \newline
- Preprocessing \newline
- Knowledge graph construction \newline
Event Log with Type level and Instance Level visualization \newline
Knowledge Graph Network visualization \newline
\end{comment}

\begin{figure}[htb]
    \centering
    \includegraphics[width=\textwidth]{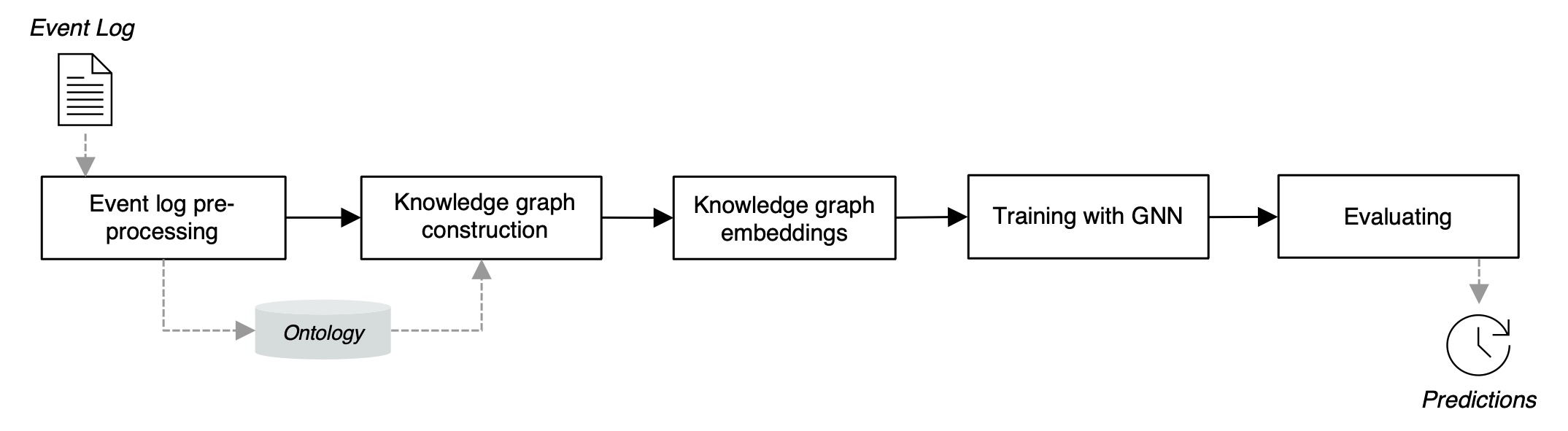}
    \caption{Pipeline of GNN4PPM}
    \label{approach}
\end{figure}

In this section, we present our graph-based approach, GNN4PPM. 
The solution predicts the next events along with the complete data payload of the event log by following the steps presented in Fig.~\ref{approach}. 
In the offline phase, an event log is first loaded, preprocessed, and transformed into a knowledge graph using an event log ontology and mapping rules. 
RDF2Vec embeddings are generated from the knowledge graph structure to capture semantic relationships between entities.
%Based on the structure of the knowledge graph, RDF2Vec graph embeddings are generated and used as an initial representation to train the R-GCN model.
These embeddings serve as initial node features for an R-GCN, which learns to predict event attributes by propagating information through the graph's type edges. 
%The R-GCN model captures relational dependencies between event log entities via typed edges.% data of the event log attributes. 
The trained R-GCN model is applied to validation cases in the online phase to generate predictions.
%The trained graph serves as input for validation cases, which can be performed in the online phase to generate predictions.
\vspace{-15pt}
\vspace{-20pt}
\subsection{Event Log Knowledge Graph Construction}
To enable relational learning over event logs, we transform the tabular event log representation into a multi-relational knowledge graph. 
This representation preserves the structural dependencies among cases, events, and attributes and enables relational message parsing in the R-GCN model.
In contrast to purely sequential encoding, the graph structure allows explicit modeling of temporal dependencies between attribute values and cross-attribute interactions.

%The first step in the pipeline is to load the event log, preprocess it, and construct a knowledge graph based on it, preserving its semantic meaning. 

\hspace*{15pt} The first step of the pipeline transforms the event log into a structured knowledge graph while preserving its semantics and temporal relations.
%\paragraph{Event preprocessing.}
During preprocessing, we extend the classical definition of an event $e$ by introducing a unique event identifier $\mathit{eid} \in \mathbb{N}^+$, reflecting the execution order within a case. Modeling $\mathit{eid}$ as an explicit node in the KG, as shown in Fig.~\ref{graph}, prevents identical activity labels across different cases from being merged and ensures that relational dependencies are captured at the level of concrete event instances. An event is thus represented as 
$e = (\mathit{eid}, \mathit{cid}, a, t, v_1, \ldots, v_n).$

\hspace*{15pt} All other components of the tuple follow Definition~\ref{def:event}.
Furthermore, the \emph{event attributes} $v_1, \ldots, v_n$, and \emph{trace attributes} $w_1, \ldots, w_m$, are classified into \emph{numerical} or \emph{categorical} to facilitate the embedding and training.
\vspace{-10pt}
\subsubsection{Event Log Knowledge Graph.}
Given an event log $L = \{\sigma_1,\ldots,\sigma_z\}$, we construct
an event log knowledge graph $\mathcal{KG}_{L}$ consisting of RDF triples.
The graph represents the structural relations between cases, events, and attribute values contained in the event log.

\hspace*{15pt} Each trace $\sigma = \langle e_1, e_2, \ldots, e_{|\sigma|} \rangle$ 
consists of events 
$e = (\mathit{eid}, \mathit{cid}, a, t, v_1, \ldots, v_n)$ 
and trace-level attributes $(\mathit{cid}, w_1, \ldots, w_m)$, as defined in Section~\ref{PRE}. The knowledge graph is constructed by generating RDF triples describing event attributes, trace relations, and temporal dependencies between events.
For each event $e$, we create triples linking the event identifier to its activity label, timestamp, and attribute values:
%we create the set of triples
%Each attribute of event $e$ and trace $\sigma$ are connected by a \emph{set of triples} $(s,r,o)$ 
%that express their relations:
\begin{align}
T_e &= \{(\mathit{eid}, r_a, a), (\mathit{eid}, r_t, t)\}
      \cup \{(\mathit{eid}, r_{v_k}, v_k) \mid 1 \le k \le n\}, 
\end{align}

%linking the event identifier to its activity, timestamp, and event-specific attribute values.
These triples connect the event node to the nodes representing its activity label and attribute values.
%\paragraph{Trace triples.}
Trace-level relations connect events to their corresponding case ID and represent trace attributes:
%Trace attributes are represented analogously:
\begin{align}
T_\sigma &= \{(\mathit{cid}, r_{w_j}, w_j) \mid 1 \le j \le m\}
            \cup \{(\mathit{cid}, r_{\text{contains}}, \mathit{eid}_i) \mid e_i \in \sigma, \ 1 \le i \le |\sigma|\},
\end{align}
These triples associate each case with its trace attributes and the events belonging to that trace.
%stating which event identifiers belong to each case.
%\paragraph{Temporal relations.}
Temporal ordering within a trace is represented through a \emph{directly-follows} relation. 
For each pair of consecutive events $e_i$ and $e_{i+1}$ in a trace $\sigma$, we add triples that connect the values of the same attribute types across both events:
%For each pair of consecutive events $e_k$ and $e_{k+1}$ in a trace $\sigma$, we add triples that connect the attribute values of $e_k$ to those of $e_{k+1}$ for every attribute type $x \in \{a, t, v_1, \ldots, v_n\}$. 
For every attribute $x \in \{a,t,v_1, \ldots, v_n\}$ we denote by $x(e_i)$ the value of attribute $x$ in event $e_i$.
The set of directly-follows triples is then formally defined as:
\begin{align}
T_{df} =
\{(x(e_i), r_{\mathrm{dF}}, x(e_{i+1}))
\mid e_i,e_{i+1} \in \sigma,\ i < |\sigma|,\ \sigma \in L,\
x \in \{a,t,v_1,\ldots,v_n\}\}.
\end{align}
\begin{comment}
\[
T_{df} =
\{(x_k, r_{\mathrm{dF}}, x_{k+1})
\mid e_k, e_{k+1} \in \sigma,\ k < |\sigma|,\ \sigma \in Log,\ 
x_k, x_{k+1} \in e_k, e_{k+1}\}.
\]
\end{comment}
This relation captures how attribute values evolve from one event to the next within the same trace.
Modeling directly-follows at the attribute level, enables capturing temporal dependencies not only between events but also between evolving attribute values. 
Unlike sequential models that receive a prefix as input, GNN4PPM encodes prefix context implicitly. The knowledge graph contains, e.g., \textit{directlyFollows} edges linking consecutive attribute values across events, so the R-GCN aggregates multi-hop neighborhood information from earlier events into the embedding  during message passing. The prediction head then operates on the resulting embedding of the node, which carries the accumulated context of the preceding trace.
This design allows learning how specific attribute combinations influence subsequent attribute values.
Additionally, type triples are added to associate each instance with its corresponding domain class.
%Furthermore, the set of \emph{directly-follows} triples expresses the temporal ordering between consecutive events within the same trace.
%For all event attributes $x \in \{a, t, v_1, \ldots, v_n\}$, such that
%$T_{df} =
%\{(x_k, r_{\text{dF}}, x_{k+1})
%\mid e_k, e_{k+1} \in \sigma,\ k < |\sigma|,\ \sigma \in Log,\ 
%x_k, x_{k+1} \in e_k, e_{k+1}\}$. 
%Each triple connects the values of the same attribute type between two successive events $e_k$ and $e_{k+1}$ within the same trace $\sigma$. Additionally, the set of \emph{type} triples connects each domain to its respective type.
The resulting event log knowledge graph is given by:
\begin{align}
\mathcal{KG}_{L} = T_e \cup T_\sigma \cup T_{df}.    
\end{align}
The knowledge graph can be interpreted as a directed labeled graph $G_{L} = (V_{L},E_{L})$ where the nodes $V_{L}$ correspond to the entities and literals appearing in the triples and the edges $E_{L}$ correspond to the triples themselves. 

\begin{figure}
    \centering
    \includegraphics[width=\textwidth]{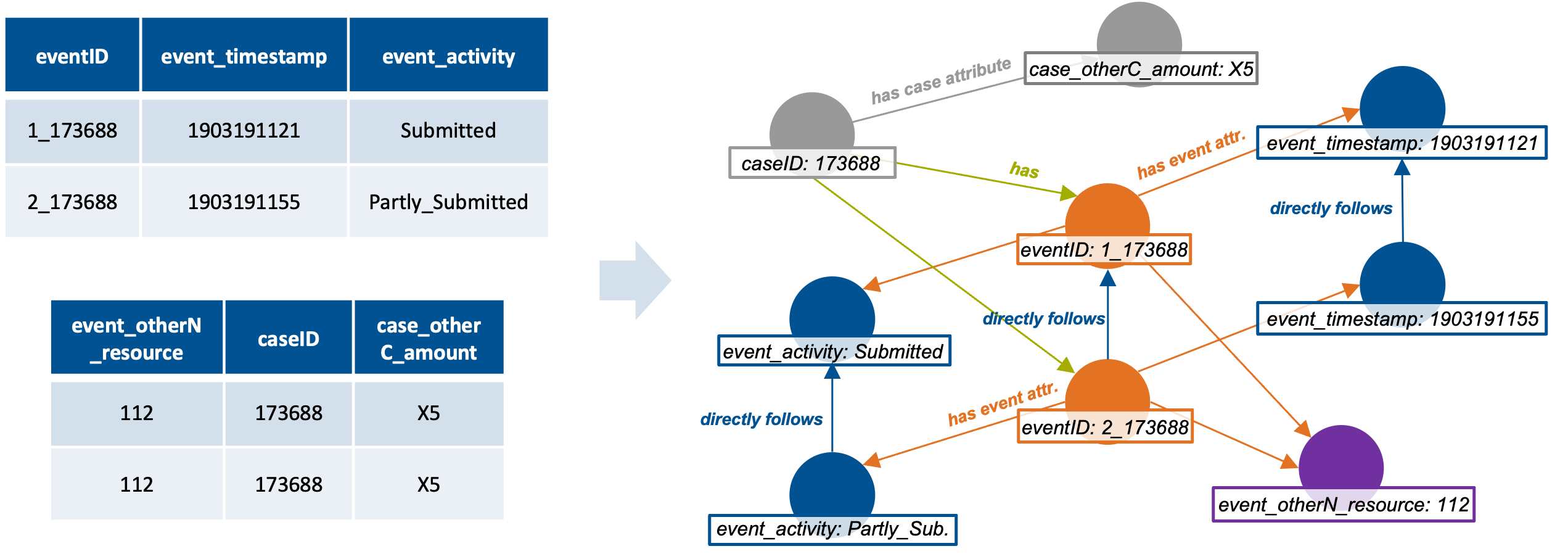}
    \caption{Knowledge graph representation of the entity relations of an event log.}
    \label{graph}
\end{figure}

\hspace*{15pt} Besides the entity relations shown in Fig.~\ref{graph}, the class relations in the event log on the left indicate that $\mathit{eid}$ contains its event attributes, and $\mathit{cid}$ contains $\mathit{eid}s$ and their case attributes. In the figure, the edges indicate the direction, and the edge labels indicate the relation type.
Additionally, all other attributes of the log were classified into \emph{other numerical} or \emph{other categorical}. 
Although their instances are unique, this classification was fundamental for the embeddings and loss calculation in the neural network. 
%To construct this \emph{Event Log Knowledge Graph}, we have identified classes, entities, and literals of the event log, as well as the corresponding relations that express their connections. 
%in a generalized form. 
The class hierarchy and conceptual relations of the event log are defined through an ontology inspired by the XES standard~\cite{guenther_xesstandard_2014}. The ontology defines node types \textit{CaseID}, \textit{EventID}, categorical attribute nodes, and numerical literals, connected by structural and temporal predicates.
Instance-level triples are generated by YARRRML-Mappings~\cite{van2021towards} and RML-Mappings~\cite{dimou2014rml}, which transfer each combination of instances of the event log into 
%These mappings serve as an efficient way of describing how each row of the event log is translated 
RDF triples, allowing the event data to be represented as a knowledge graph. Since a detailed explanation of the ontology is not the purpose of this work, a formalization of the ontological structure is left for future work.

\hspace*{15pt} The resulting knowledge graph serves as the structured input for embedding generation.
%The relations between the classes are defined in an ontology constructed based on the XES standard definition \cite{guenther_xesstandard_2014}, and the relations between the entities are defined in the RDF mappings and represented in Figure~\ref{graph}. 
%After the knowledge graph is constructed using RDF-Mappings~\cite{van2021towards,dimou2014rml}, the sequential information of traces and events is embedded.
Moreover, the constructed graph can be related to object-centric settings, in which object types correspond to attribute types with their own relational structure. Both approaches make use of heterogeneous graphs, which are well-suited for expressing inter-dependencies between objects.
\vspace{-10pt}
\subsection{Knowledge Graph Embeddings} 

Knowledge graph embeddings are created using RDF2Vec, which learns low-dimensional representations of entities by performing biased random walks over the RDF graph and training a Word2Vec model on the resulting sequences \cite{ristiski_RDF2Vec_2016}. 
This embedding technique was chosen because it captures multiple paths from each entity, encoding rich contextual information from the neighbouring nodes and predicates. 
Since the learned embeddings preserve similar entities close to each other \cite{maekelburg_mperl_2025}, event attributes for each event instance can be kept closer in the vector space, making RDF2Vec a suitable embedding for the given prediction task. 

\hspace*{15pt} Performing biased walks instead of random walks allows 
%To perform biased walks in the RDF, we use Node2Vec, allowing 
us to prioritize event attributes and directly follow relations according to breadth-first and depth-first parameters \cite{grover2016node2vec}. This is important when representing our event logs since uniform random walks would pick any neighbor with equal probability without following the temporal sequence of traces.
After the biased walks are performed on the graph, we train the resulting sequences with Continuous Bag of Words (CBOW) to predict the center token from its surrounding context window \cite{mikolov2013efficient}. 
For example, in a walk such as
[\textit{caseID:173688}, \textit{has}, \textit{eventID:1}, \textit{directly follows}, \textit{eventID:2}], CBOW learns \textit{eventID:1} from the context of both its case and its next event simultaneously (see Fig. 2).

\hspace*{15pt} Since the embedding quality depends on walk and CBOW settings (e.g., walk length, window size, embedding size), we perform unsupervised model selection by clustering the embeddings with K-Means and choosing hyperparameters that maximize the silhouette score on the training set, so that similar entities land close together in the embedding space. The silhouette-based selection is performed exclusively on training-set entities, as both the random walks and the Word2Vec training are restricted to the training graph, ensuring that no information from validation cases influences the embedding hyperparameter choice. These pre-trained embeddings are stored with a corresponding \emph{id} mapping and serve as initial node features for training with the R-GCN.

\vspace{-10pt}
\subsection{Training with R-GCN}
This section describes the R-GCN architecture used for representation learning and the multi-target objective employed for model training.
\begin{comment} 
\textbf{Model Architecture.}
The model jointly predicts all target event attributes for each event-to-event transition.
Categorical attributes are predicted via independent multi-layer classification heads, while numerical attributes (including inter-event time delta) are predicted through multi-layer regression heads.
Each prediction head consists of a hidden layer with ReLU activation and dropout.
The encoder is a three-layer R-GCN with layer normalization, residual connections, and ReLU activations.
Separate training and validation graphs were constructed to preserve case-level isolation during inference.
\end{comment}
\vspace{-10pt}
\subsubsection{Model Architecture.}
The RDF2Vec entity embeddings serve as initial node features for the R-GCN encoder.
During training, R-GCN receives these embeddings along with the graph structure (edge index and edge types).
We use R-GCN because the event log knowledge graph contains multiple semantic relation types, and R-GCN explicitly models relation-specific message passing in heterogeneous graphs. Unlike conventional heterogeneous GNN frameworks that define distinct aggregation operators (e.g., GraphSAGE or GAT) per relation type, our model applies a uniform aggregation function across all edge types. Relation-type heterogeneity is captured through the relation-specific transformation matrices, which are parameterized via basis decomposition to keep the number of learnable parameters controlled across the large number of RDF predicate types in the knowledge graph.
Architectures without explicit relation modeling would treat all edges uniformly and cannot distinguish between temporal and contextual dependencies. 
%We use R-GCN because our knowledge graph contains multiple semantic relations between entities, and R-GCN is designed to handle such heterogeneous graphs where each node aggregates messages from its neighbors according to the type of relation that connects them.

\hspace*{15pt} The encoder consists of three R-GCN layers with layer normalization, residual connections, and ReLU activations, $\phi$. Three layers were chosen to allow sufficient neighborhood aggregation depth, capturing event, case, and attribute-level context. A residual connection from the first to the second layer, combined with layer normalization after each of the first two layers, preserves node-specific information and stabilizes training, mitigating the risk of over-smoothing.
In the forward pass, each R-GCN layer aggregates messages from neighboring nodes, with separate transformation matrices for each relation type. 
To handle the multiple relation types in our event log knowledge graphs, we employ basis decomposition, which represents relation-specific weight matrices as linear combinations of a smaller set of basis matrices.
%That means, gradients flow back through these weight matrices during training. 
%We use this transformation since it has been shown effective at accumulating and encoding features from neighborhoods \cite{schlichtkrull2018rgcn}, and our graph accumulates several neighborhood features through the relations between the \emph{event ID} and its attributes. 
Formally, the forward propagation of each R-GCN layer follows the update rule proposed by \cite{schlichtkrull2018rgcn}. For a node \(i\) at layer \(l\), the updated representation \(h^{(l+1)}_{i}\) is computed as:

\begin{equation}
h^{(l+1)}_i = \phi\!\left(
    \sum_{r \in \mathcal{R}} 
    \sum_{j \in \mathcal{N}^r_i}
        \frac{1}{c_{i,r}} 
        W^{(l)}_{r} h^{(l)}_j
    \;+\;
    W^{(l)}_0 h^{(l)}_i
\right),
\label{eq:rgcn_forward}
\end{equation}

where \(\mathcal{N}^r_i\) is the set of neighbors of \(i\) under relation \(r\),
\(W^{(l)}_{r}\) is the relation-specific transformation matrix, 
\(W^{(l)}_0\) is the self-loop weight, $\phi(\cdot)$ is the ReLU activation function, and $c_{i,r}=|\mathcal{N}^r_i|$ is a degree-based normalization constant equal to the number of neighbors connected to node $i$ through relation $r$.
%This formulation aggregates transformed neighbor features through a normalized sum and contains relation-specific transformations for edge type and direction, which makes, for example, the directly follows relations of the graph more meaningful during propagation.
This formulation enables the model to learn different aggregation patterns for different relation types, making relations such as \emph{directly-follows} or \emph{has-resource} more meaningful during information propagation.

%The model uses the structure of the knowledge graph created from the event log to compute hidden features, which are calculated according to the GNN architecture. 

%The GNN propagates the information through the graph structure and updates using the functions of PyTorch Geometric (take off this last). Why we choose RDF2Vec. Add a figure of RDF2Vec EMbbeddings. 
% Write about the knowledge graph embeddings more details

%GNN Architecture, explain why we use r-gcn ? what they are doing

%This is the Forward pass being used.
%Each training sample consists of a prefix sequence (input) and a suffix sequence (target) (Figure~\ref{sequences}), with all attributes of the event log preserved. 
\vspace{-16pt}
\subsubsection{Training Objective.}
%Loss:
%I need to change the variables to not use the same as defined before
The model is trained in a multi-target setting, predicting all categorical and numerical attributes of the next event.
%Finally, the loss function to train the model is calculated by predicting all categorical and numerical attributes at every time step. 
Since these attributes differ fundamentally in their type and scale, a unified loss function would not adequately capture their prediction objectives; therefore, multiple loss functions are required.  
%Multiple loss functions are necessary because categorical and numerical differ in their output space and statistical structure.
%Categorical attributes lie in a discrete label space and require probabilistic modeling, whereas numerical attributes lie in a continuous space and require regression objectives. 
%Cross-entropy models discrete probability distributions, whereas MSE and Huber loss optimize regression objectives in continuous space.

%\paragraph{Categorical Attributes}
\hspace*{15pt} For categorical attributes, we use cross-entropy loss with inverse frequency class weighting to address class imbalance~\cite{he2009learning}. 
Cross-entropy is well-suited for multi-class prediction, as it directly optimizes the likelihood of the correct class while producing calibrated probability estimates. 
Event logs typically have an unbalanced attribute value distribution. 
Inverse frequency weighting compensates these imbalances and prevents the model from biased predictions. 
For a categorical attribute $k$, the loss, $\mathcal{L}_{\mathrm{CE}}$, is defined as:
\begin{equation}
\mathcal{L}_{\mathrm{CE}}^{(k)} 
= \mathrm{CE}\!\left( \hat{y}^{(k)}, y^{(k)}, w_k \right),
\end{equation}
where $w_k$ denotes the inverse-frequency class weights.
%\paragraph{Timestamp}
For the timestamp, we apply the mean squared error (MSE) on log-transformed and normalized values.
Time differences in real-world processes are often unbalanced. Therefore, applying a logarithmic transformation compresses large time gaps and prevents them from dominating during training~\cite{osborne2002notes}.  
After normalization, we define the timestamp loss $\mathcal{L}_t$ using MSE, which provides a simple and effective regression objective for continuous time prediction, since the values are now distributed more balanced.
%The timestamp loss is defined as
%\begin{equation}
%\mathcal{L}_{t} 
%= \mathrm{MSE}\!\left( \hat{y}^{(t)}, y^{(t)} \right).
%\end{equation}

%\paragraph{Numerical Attributes.}
\hspace*{15pt} For other numerical attributes, we employ Huber Loss~\cite{huber1992robust} to improve robustness against outliers. 
%For a numerical attribute $k$, the loss is defined as
%\begin{equation}
%\mathcal{L}_{\mathrm{Huber}}^{(k)} 
%= \mathrm{Huber}\!\left( \hat{y}^{(k)}, y^{(k)} \right).
%\end{equation}
For a numerical attribute $k$, the corresponding loss $\mathcal{L}_{\mathrm{Huber}}^{(k)}$ is defined as:
\begin{equation}
\mathcal{L}_{t} = \mathrm{MSE}\!\left( \hat{y}^{(t)}, y^{(t)} \right),
\qquad
\mathcal{L}_{\mathrm{Huber}}^{(k)} = \mathrm{Huber}\!\left( \hat{y}^{(k)}, y^{(k)} \right).
\end{equation}

Unknown or missing target values are masked and excluded from gradient computation. 
%\paragraph{Overall Training Objective.}
The total loss is computed as a weighted aggregation of the individual task losses:
\begin{equation}
\mathcal{L}_{\text{total}}
=
\mathcal{L}_{\mathrm{CE}}^{(\mathrm{act})}
+
\lambda_t \, \mathcal{L}_t
+
\lambda_c \sum_{k \in \mathcal{C}} \mathcal{L}_{\mathrm{CE}}^{(k)}
+
\lambda_n \sum_{k \in \mathcal{N}} \mathcal{L}_{\mathrm{Huber}}^{(k)},
\label{eq:multitaskloss}
\end{equation}
where $\mathcal{C}$ and $\mathcal{N}$ denote the sets of categorical and numerical attributes, respectively,
and $\lambda_t$, $\lambda_c$, and $\lambda_n$ are task-specific balancing coefficients optimized via hyperparameter search.
\begin{comment}
For each attribute $k$, we assign a weight $w_k$. Given predictions $\hat{y}^{(k)}_{b,l}$ and targets $y^{(k)}_{b,l}$ for batch index $b$ and sequence position $l$, we do not count the padding positions in the loss since they were added to match the same length, therefore the mask $ m_{b,l} = \mathbf{1}[\, y^{(\mathrm{act})}_{b,l} \neq \mathrm{PAD} \,]$. The loss is the weighted sum of the Cross-entropy (CE) and masked mean-squared error (MSE).
% Maybe not necessary the masked padding information, check

\begin{equation}
\mathcal{L}_{\text{total}}
= \sum_{k \in \mathcal{C}} 
w_k \, \mathrm{CE}\!\left( \hat{y}^{(k)}, y^{(k)} \right)
\;+\;
\sum_{k \in \mathcal{T}} 
w_k \,
\frac{
\sum_{b,l} m_{b,l} \left( \hat{y}^{(k)}_{b,l} - y^{(k)}_{b,l} \right)^{2}
}{
\sum_{b,l} m_{b,l} + \varepsilon
}.
\label{eq:multitaskloss}
\end{equation}
    
\end{comment}
% Why two loss functions?
% Why is one better fitting for each type?
% Decoding?
%Two different loss functions are used because categorical and numerical attributes have fundamentally different output spaces and statistical properties. With CE, the model outputs a probability distribution over classes, and with masked MSE, it minimizes the \(\ell_2\) distance between prediction and target for continuous values.

%.Decoder

\begin{comment}
- Knowledge Graph Embeddings \newline
(TransE, RDF2Vec) \newline
- Graph Neural Network \newline
(GAT, [GCN or R-GCN]) \newline
(Decoder) \newline
- Prediction Task
(Next Activity Prediction, Next Event and its attributes)
\end{comment}

\section{Evaluation}\label{EVA}

%In this section, we describe the event log and the experimental setup to evaluate the performance of GNN4PPM, as well as the implementation, results, and comparison with state-of-the-art solutions. 
In this section, we evaluate the performance of GNN on several real-world event logs. 
Our evaluation addresses: (Q1)
%\begin{enumerate}[label=(Q\arabic*)]
How does GNN4PPM perform on the next activity prediction task compared to the state-of-the-art methods; %(\S\ref{sec:SOTA})
(Q2) How accurately can the model predict the additional attributes of the next event; and %(\S\ref{sec:multiAttribut})
(Q3) How does each event attribute influence the prediction performance. %(\S\ref{sec:attributInfluence})
%\end{enumerate}
We first describe the datasets and experimental setup, then compare with state-of-the-art approaches.
We then analyze the results of our defined multi-target task, and lastly, investigate the influence of individual event attributes on the prediction performance. The source code is available online\footnote{\url{https://github.com/ana-luisa-costa/GNN4PPM.git}}.   

\subsection{Datasets}
We used six real-life event logs to evaluate our solution (see Table~\ref{tab:combined_characteristics}). All used logs are publicly available at 4TU Centre for Research~\cite{4tu_researchdata} and contain processes performed in different domains. The datasets BPIC12A, BPIC12W, and BPIC12WC describe the application process for a personal loan or overdraft at a Dutch Financial Institute. The first contains the complete lifecycle traces of the subprocess 'Application', the second contains the subprocess 'Work', and the third contains the complete lifecycle traces of the subprocess 'Work'. The event log BPIC13O consists of the open problems in the problem management process at VolvoIT Belgium, which includes the activities required to diagnose the root cause of incidents and to secure the resolution of those problems. While BPIC20P contains events pertaining to two years of travel expense claims, BPIC20R contains events related to the request for payments.

\begin{table*}[t]
\centering
\renewcommand{\arraystretch}{1.2}
\caption{Structural characteristics of the event logs and constructed knowledge graph information.}
\label{tab:combined_characteristics}
\resizebox{\textwidth}{!}{%
\rowcolors{2}{gray!15}{white}
\begin{tabular}{lcccccccc}
\toprule
\rowcolor{gray!25}
\textbf{Dataset} 
& \textbf{\#Traces} 
& \textbf{\#Events} 
& \textbf{\#Variants}
& \textbf{\#Case attr.}
& \textbf{\#Event attr.}
& \textbf{Triples}
& \textbf{Entities}
& \textbf{Predicates} \\
\midrule

BPIC12A  & 13{,}087 & 60{,}849 & 17 & 1 & 3 & 597{,}071  & 144{,}870  & 10 \\
BPIC12W  & 9{,}658 & 170{,}107 & 2921 & 1 & 3 & 1{,}554{,}389 & 360{,}122  & 10 \\
BPIC12WC & 9{,}658 & 72{,}413 & 2963 & 1 & 3 & 685{,}039 & 164{,}741 & 10 \\
BPIC13O  & 819 & 2{,}351 & 108 & 3 & 6 & 32{,}925 & 5{,}985 & 14 \\
BPIC20P  & 10{,}500 & 56{,}437 & 896 & 8 & 5 & 254{,}401 & 59{,}081  & 16 \\
BPIC20R  & 6{,}886 & 36{,}796 & 76 & 3 & 4 & 368{,}206 & 110{,}655  & 7 \\

\bottomrule
\end{tabular}
}
\end{table*}
\vspace{-10pt}
\subsection{Experimental Setup}
This section describes the experimental configuration, including data handling, training procedure, hyperparameter optimization and evaluation.
%The approach was implemented in Python using PyTorch and PyTorch Geometric.
%RDF2Vec embeddings were generated using the pyrdf2vec library based on Gensim.
All experiments were conducted on a compute node using one NVIDIA H100 GPU with 96 GB of memory.
%All experiments were conducted on a machine equipped with an Apple M4 Pro chip and 24 GB RAM.
%a compute node equipped with two Intel Xeon Gold 6548N processors, 2 TB of RAM, 15 TB of NVMe SSD storage, and four NVIDIA H100 GPUs with 96 GB of memory each. 

\textbf{Data Handling and Splits.}
The prediction task is formulated as event-to-event prediction.
For each consecutive pair $(e_i, e_{i+1})$ within a trace, the model uses the graph representation at event $e_i$ to predict the attributes of $e_{i+1}$.
Data were split at the case level (80\% training, 20\% validation) using a fixed random seed, following the random case-level split convention used in benchmark studies in this domain \cite{RamaManeiro2022}.
All embeddings and graph structures were constructed exclusively from training cases.  

\textbf{Hyperparameter Optimization.}
Model hyperparameters were selected using automated Bayesian optimization with 15 trials and early pruning to discard underperforming configurations.
The search space included: hidden dimension $\{64, 128, 256\}$, prediction head hidden dimension $\{64, 128, 256\}$, number of basis functions $[5, 15]$, dropout rate $[0.1, 0.5]$, learning rate $[10^{-4}, 10^{-2}]$ (log scale), and task-specific loss weights for multi-task balancing.
Each trial was trained for $100$ epochs, and the configuration with the lowest validation loss was selected for final training. The task-specific loss balancing coefficients were optimized over the following search ranges: $\lambda_t \in [0.1,1.0]$, $\lambda_c \in [0.3,1.0]$, and $\lambda_n \in [0.1,0.5]$.

\textbf{RDF2Vec Configuration.} 
The walk length, CBOW context window size, and embedding dimensionality were selected through unsupervised Bayesian optimization, maximizing the silhouette score on K-Means training embeddings. The corresponding search spaces were $[6,12]$ for walk length, $[3,8]$ for context window size, and $\{64,128,256\}$ for embedding dimensionality.

\textbf{Training Configuration.}
Training was performed using the Adam optimizer with cosine annealing.
We applied early stopping with a patience of 30 epochs, gradient clipping with a maximum norm of $1.0$. 
Models were trained for up to $100$ epochs.
%Inverse frequency weighting was applied to categorical attributes to address class imbalance.
%The timestamp prediction target was the log-transformed inter-event time delta (in seconds), normalized using training set statistics.

\textbf{Evaluation Metrics.}
Categorical attributes are evaluated using macro-averaged precision, recall, F1-score, and accuracy.
Numerical attributes are evaluated using Mean Absolute Error (MAE) in normalized log-space.
%Numerical attributes were predicted via regression on log-normalized deltas and evaluated using Mean Absolute Error (MAE) in normalized log-space.
All reported metrics were computed exclusively on the validation set.
\vspace{-10pt}
\subsection{Comparison against State-of-the-Art}
\label{sec:SOTA}
%\jm{framing that other approaches only do the next activity prediction, while we do all other accuracies}
To put our performance into perspective and address (Q1), we compare our approach against a set of state-of-the-art methods.
%~\cite{Dissegna2025,pasquadibisceglie_lupin_2024,pasquadibisceglie_prophet_2024,Rauch2026,Camargo2019,Evermann2017,tax_predictive_2017,Theis2019} 
All compared approaches focus on the NAP task since they do not allow for the prediction of the other attributes in the same training. In contrast, our approach predicts the next event together with its associated attributes. However, to ensure comparability with existing work, we report NAP accuracy in this section.

\hspace*{15pt} The most recent approaches \cite{Dissegna2025,pasquadibisceglie_lupin_2024,pasquadibisceglie_prophet_2024,Rauch2026} were implemented and executed by us for all datasets. The older approaches \cite{Camargo2019,Evermann2017,tax_predictive_2017,Theis2019} were run by us for datasets BPIC13O, BPIC20P, and BPIC20R. For the other three datasets, we rely on the results of \cite{RamaManeiro2022}, who performed an extensive benchmark on these logs. 
For each method, we followed the training configuration and evaluation protocol recommended in the respective publication, including the specified data splits, cross-validation procedures, and, where available, hyperparameter settings. For datasets on which certain approaches were not originally evaluated, we applied the same configuration strategy proposed by the authors and conducted additional experiments accordingly.
In these cases, hyperparameters were selected according to published guidelines or the default settings provided in the official implementations. Table~\ref{tab:sota_accuracy} summarizes the accuracy results across all benchmark datasets. Other measures such as F1 score, precision, and recall, are available in the data file online\footnote{\url{https://figshare.com/s/cde2c35c6ab3f7ca5422}}.

\begin{table*}[t]
\centering
\renewcommand{\arraystretch}{1.2}
\caption{Accuracy of the Next Activity Prediction Task compared to different approaches. 
Best result per dataset highlighted in bold; second-best is underlined.}
\label{tab:sota_accuracy}

\resizebox{\textwidth}{!}{%
\rowcolors{2}{gray!15}{white}
\begin{tabular}{lcccccc}
\toprule
\rowcolor{gray!25}
\textbf{Approach} 
& \textbf{BPIC12A} 
& \textbf{BPIC12W} 
& \textbf{BPIC12WC} 
& \textbf{BPIC13O} 
& \textbf{BPIC20P} 
& \textbf{BPIC20R} \\
\midrule
SEPHIGRAPH \cite{Dissegna2025}  & 67.8 & 78.6 & 46.6 & 20.2 & 73.8 & 76.5 \\
LUPIN \cite{pasquadibisceglie_lupin_2024}      & 19.3 & 35.7 & 21.2 & 24.4 & 39.7 & 51.1 \\
PROPHET \cite{pasquadibisceglie_prophet_2024}    & 76.7 & \textbf{89.2} & \underline{80.3} & 62.0 & 83.8 & 87.8 \\
BEST \cite{Rauch2026}       & 76.1 & 69.0 & 40.0 & 48.8 & \underline{86.9} & 89.0 \\
CAMARGO \cite{Camargo2019}    & 75.9 & 76.4 & 68.9 & \underline{77.2} & 84.4 & \textbf{91.9} \\
EVERMANN \cite{Evermann2017}   & 75.8 & 75.4 & 67.5 & 65.1 & \textbf{88.9} & \underline{90.5} \\
TAX \cite{tax_predictive_2017}        & \underline{79.5} & 85.3 & 69.8 & 60.9 & 84.4 & 89.0 \\
THEIS \cite{Theis2019}      & 65.5 & \underline{86.2} & 80.1 & 59.3 & 83.2 & 89.7 \\
\hline
\multicolumn{7}{c}{\textit{Our approach}} \\ \hline
GNN4PPM     & \textbf{88.8} & 80.8 & \textbf{83.2} & \textbf{78.1} & 80.4 & 87.8 \\
\bottomrule
\end{tabular}
}
\end{table*}

\hspace*{15pt} Overall, GNN4PPM achieves the best performance on BPIC12A, BPIC12WC, and BPIC13O, and remains competitive on the remaining three. 
BPIC13O is the smallest dataset in terms of traces and events, but it contains a relatively high number of case and event attributes relative to its size. On this dataset, GNN4PPM outperforms all competing methods. Both BPIC12A and BPIC12WC, while larger, exhibit a moderate number of event attributes and a high number of traces. GNN4PPM also achieves the strongest performance on these datasets, indicating that the approach performs particularly well on small to medium-sized event logs. In contrast, BPIC12W contains more events and a low number of event attributes.
On this dataset, PROPHET achieves the best performance. 
%For BPIC12WC, which is derived from the same source but contains fewer events and activities, the performance gap between PROPHET and GNN4PPM becomes small. 
This suggests that while certain approaches may benefit from very large logs, our approach maintains stable performance across different scales. BPIC20P and BPIC20R contain a larger number of activities and more case and event attributes. On these datasets, Evermann and Camargo achieve the strongest results, while GNN4PPM remains competitive, especially on BPIC20R. 

\hspace*{15pt} The results show that GNN4PPM performs well across different log sizes with multiple event attributes, while remaining competitive on larger, more complex datasets. These results demonstrate that predicting the complete data payload positively affects activity prediction performance while achieving competitive results across all datasets.
\vspace{-10pt}
\subsection{Multi-Target Prediction}
\label{sec:multiAttribut}
\begin{table*}[b]
\centering
\renewcommand{\arraystretch}{1.2}
\caption{Multi-target prediction results across datasets given as accuracy for event attributes and as normalized MAE for timestamp.}
\label{tab:prediction_results}
\scriptsize
\resizebox{\textwidth}{!}{%
\rowcolors{2}{gray!15}{white}
\begin{tabular}{lcccccccc}

\specialrule{0.08em}{0pt}{0pt} % top rule, no white padding

\rowcolor{gray!15}
\textbf{Dataset}
& \textbf{activity}
& \textbf{resourcecountry}
& \textbf{org\_resource}
& \textbf{lifecycle}
& \textbf{org\_group}
& \textbf{org\_role}
& \textbf{timestamp (MAE)} \\

\specialrule{0.05em}{0pt}{0pt} % rule under header, no white padding

BPIC12A  & 88.75 & 64.20 & 100 & -- & -- & -- & 0.3184 \\
BPIC12W  & 80.34 & 56.76 & 83.46 & -- & -- & -- & 0.5476\\
BPIC12WC & 83.19 & 14.83 & 100 & -- & -- & -- & 0.7667\\
BPIC13O  & 78.15 & 63.16 & 67.04 & 97.78 & 87.14 & 91.85 & 0.7100 \\
BPIC20P  & 80.44 & 99.38 & -- & -- & 88.67 & -- & 0.3785 \\
BPIC20R  & 87.79 & 99.58 & -- & -- & 90.53 & -- & 0.6400 \\

\specialrule{0.08em}{0pt}{0pt} % bottom rule, no white padding

\end{tabular}
}
\end{table*}

To address the question (Q2) and the problem definition in Section~\ref{def:problem}, our approach predicts the next event together with its complete data payload. Fig.~\ref{fig:attributes} shows the number of distinct attributes per dataset and Table~\ref{tab:prediction_results} summarizes the prediction results across all datasets. Since not all datasets contain the same event attributes, some entries are not applicable.

\hspace*{15pt} For the \texttt{timestamp} prediction results  the prediction with lowest errors are observed for BPIC12A (0.3184) and BPIC20P (0.3785), while BPIC13O (0.7100) and BPIC12WC (0.7667) have relatively large errors. This can be due to how regular the intervals between timestamp is in each process. While BPIC12A and BPIC20P have more consistent intervals between events, BPIC13O and BPIC12WC have times that range from seconds to days, making them harder to predict. Another influencing factor is the number of targets in the prediction task, as BPIC13O has the most categorical prediction heads, so the encoder must balance multiple classification gradients.

\begin{minipage}[t]{0.55\textwidth}
    \vspace{0pt}
    \centering
    \includegraphics[width=\textwidth]{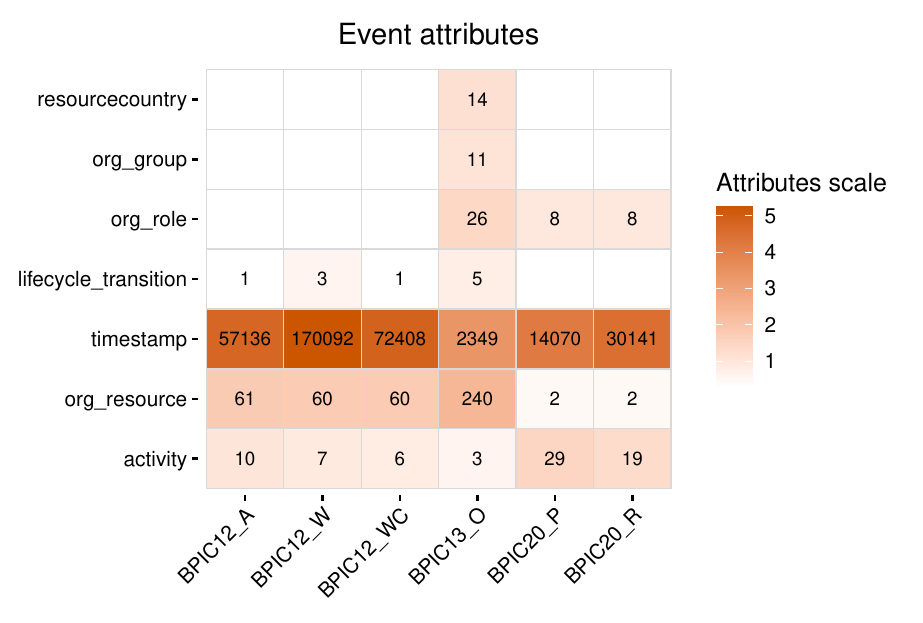}
    \captionof{figure}{Number of distinct event attributes}
    \label{fig:attributes}
\end{minipage}
\begin{minipage}[t]{0.44\textwidth}
\vspace{0pt}

For predictions in which attributes are classes, the number of distinct values highly influences the results. In datasets like BPIC20P and BPIC20R, where the number of distinct \texttt{org\_resource}-values is low, the accuracy is very high (99.38\% and 99.56\% respectively), in comparison to datasets with many distinct resources. The same effect is observed in the prediction of the \texttt{lifecycle} and \texttt{org\_role} attributes. BPIC13O shows the lowest accuracy among the
\end{minipage}

\vspace{8pt}
 datasets, but also has the most distinct values for these attributes. Overall, the results indicate that the prediction difficulty of individual attributes is influenced by attribute cardinality, the number of distinct values per class, and consistency on intervals between events, indicating that datasets with lower concept drift would perform better in the multi-target task. Attributes with few distinct identifiers can be predicted reliably, whereas those with very large numbers, such as resources or role, remain challenging. %Furthermore, the multi-target event prediction can naturally handle the prediction in object-centric logs, where event attributes reference multiple objects of different types, and case attributes are analogous to global influences not captured through direct object-to-object relations.

\subsection{Attribute Influence}
\label{sec:attributInfluence}
\begin{figure}[!h]
    \centering
    \includegraphics[width=\textwidth]{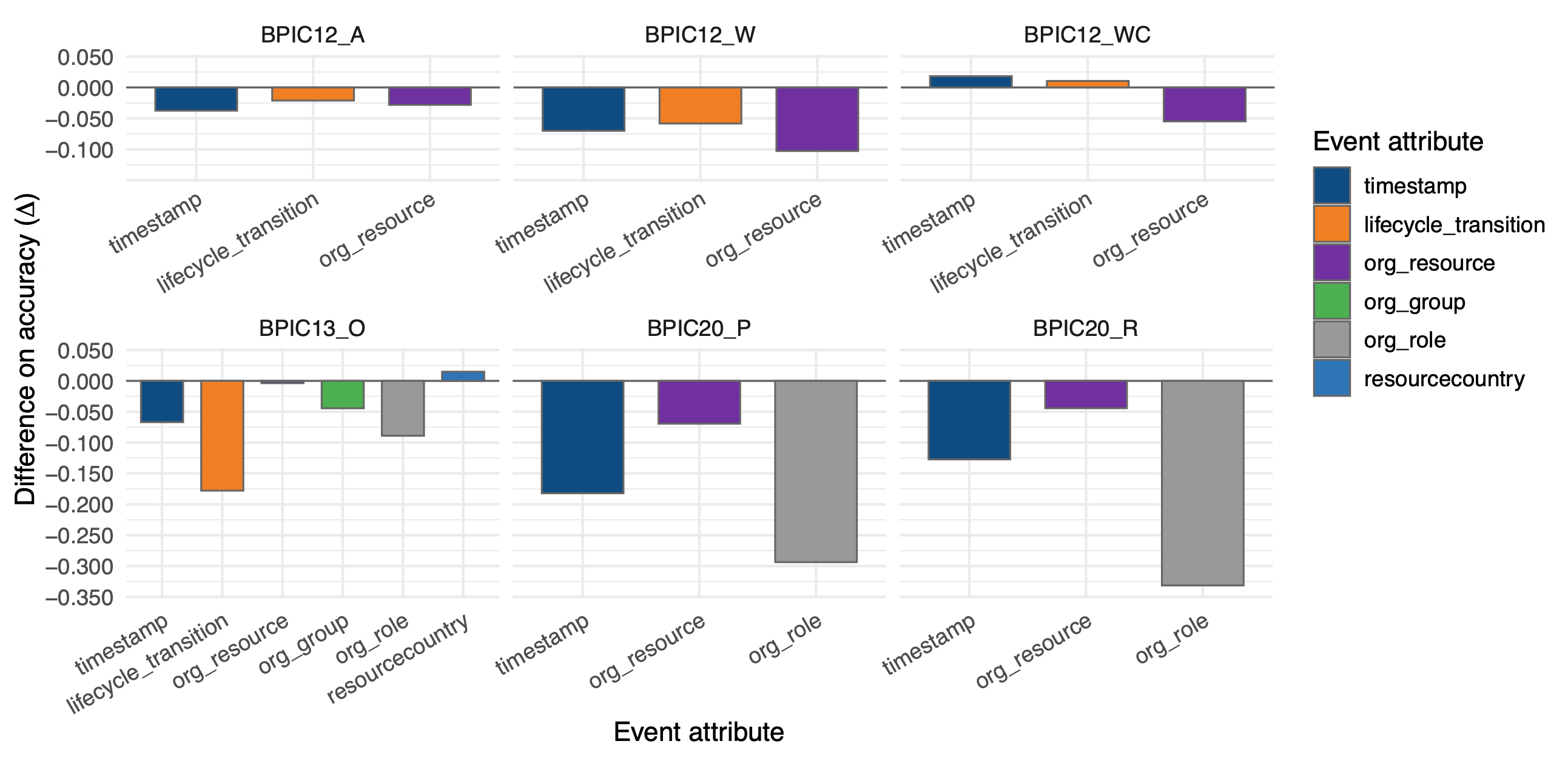}
    \caption{Influence of each event attribute on activity prediction accuracy when removed from the model}
    \label{fig:attribute_influence}
\end{figure}

To address the question (Q3), we analyze how individual event attributes influence the prediction performance.
Fig.~\ref{fig:attribute_influence} illustrates the influence of individual event attributes on the activity prediction accuracy across the datasets.
The values ($\Delta$) represent the difference in accuracy when a specific attribute is not in the model. For the comparison, the attribute was removed from the raw dataset, and the model was retrained with the new setting. Negative values, therefore, indicate that the attribute contributes positively to prediction performance, while positive values indicate that removing the attribute improves accuracy.

\hspace*{15pt} Across most datasets, the \texttt{timestamp} attribute positively influences prediction accuracy. 
Removing it leads to a decrease in performance, particularly in BPIC20P ($\Delta=-0.1821$) and BPIC20R ($\Delta=-0.1272$).
%Although timestamps exhibit a large number of distinct values (see Figure \ref{fig:attributes}), they capture the temporal progression of the process and therefore provide important predictive signals. 
The \texttt{org\_resource} and \texttt{org\_role} attributes also improves prediction performance. For example, removing the first reduces accuracy in BPIC12W ($\Delta=-0.1027$) and BPIC20P ($\Delta=-0.0694$), while removing the second results in large decreases in accuracy ($\Delta=-0.2937$ and $\Delta=-0.3314$). %indicating that resource assignments provide useful contextual information about process execution.
%The strongest influence is observed for the \textit{organization role} attribute in the BPIC20 datasets, where removing this attribute results in large decreases in accuracy ($\Delta=-0.2937$ and $\Delta=-0.3314$). 
%This suggests that the organizational role executing an activity is highly indicative of the next process step.
At the same time, some attributes show small positive values, meaning that removing them slightly improves prediction performance. This effect appears, for example, in BPIC12WC, where removing \texttt{timestamp} ($\Delta=0.0182$) or \texttt{lifecycle\_transition} ($\Delta=0.0105$) slightly increases accuracy. These attributes either have high cardinality or weak correlations with the control flow, which may introduce noise into the model.

\hspace*{15pt} The results indicate that temporal information, resources, and organizational roles provide the strongest predictive targets, while some higher-level attributes may contribute little information.
Therefore, including case and event attributes as context in the knowledge graph improves the prediction performance of the next activity prediction task, achieving better accuracy across many datasets when compared with an architecture that considers fewer attributes. 
%while remaining robust to attributes that contribute little predictive information.
%These results highlight that explicitly modeling event attributes in the knowledge graph allows the model to capture which contextual factors are most informative for predicting the next process step.

%\subsection{Limitations and Future Work}

\section{Conclusion}

We introduce GNN4PPM, an approach to multi-target predictions that captures contextual relations among event and case attributes that sequential models do not preserve.
%We introduce GNN4PPM, a graph-based approach for multi-target predictive process monitoring that considers the full payload of the next event in a running process instance. By combining a heterogeneous knowledge graph, RDF2Vec embeddings, and an R-GCN architecture, the approach captures contextual relations among event and case attributes that are not preserved by sequential models. Experimental results demonstrate competitive predictive performance across multiple event logs.
%that captures temporal dependencies between attribute values and cross-attribute interactions that sequential models do not preserve.
%, and case attributes are global influences that are not explicitly modeled through object interactions.
Despite these promising results, several limitations must be acknowledged. Experiments showed that %that extending the prediction target to the full data payload improves the prediction accuracy of a specific attribute. 
%such as the next activity prediction.
%Further influencing factors are the attribute cardinality and the regularity of time intervals.
%This demonstrates that graph-based, multi-target approaches are an alternative to the single-target paradigm and do not come at the cost of prediction accuracy, but rather enhance it.
GNN4PPM performs well on small to medium-sized event logs but is less competitive on larger logs that contain fewer attributes, such as the BPIC12W. 
Furthermore, prediction accuracy is directly related to attribute cardinality, since attributes with many distinct values, such as resources and roles, remain difficult to predict regardless of the model architecture. The timestamp prediction is also sensitive to the process irregularity of time intervals. Additionally, a direct baseline comparison for the multi-target task is not possible since other approaches do not predict the complete data payload of the next event. Finally, the evaluation is further limited by the absence of an independent test set and by being conducted on six publicly available event logs, all drawn from a limited set of domains, which restricts the generalization of the findings to other processes where attribute structures differ.

\hspace*{15pt} Future work includes first extending GNN4PPM from next-event to suffix prediction, which allows predicting complete sequences of future events with their full data payloads. Second, there is a potential for applying the approach to the prediction of objects in object-centric logs. %since the knowledge graph structure naturally encodes multiple typed relations. 
Our heterogeneous graph structure, which is better suited for expressing inter-dependencies between attributes, indicates the applicability of GNN4PPM to object-centric event logs, where event attributes are objects with different type relations in the knowledge graph. 
Third, formalizing in detail the event log knowledge graph ontology and schema and automating the graph construction.
Additionally, incorporating explainability into the R-GCN predictions, for example, through graph attention or post-hoc explanation methods, would allow identifying which attribute relations influence individual predictions. Lastly, evaluating GNN4PPM on a broader set of domains, which include multiple targets to be predicted, would strengthen the generalization of the findings.

\label{CON}

%
% ---- Bibliography ----
%
% BibTeX users should specify bibliography style 'splncs04'.
% References will then be sorted and formatted in the correct style.
%
\bibliographystyle{splncs04}
\bibliography{references.bib}

\end{document}